\documentclass[letter, 10pt, conference]{ieeeconf}
\IEEEoverridecommandlockouts

\usepackage{cite}

\usepackage{amsmath,amssymb,amsfonts}
\usepackage{algorithmic}
\usepackage{graphicx}
\usepackage{subcaption}
\usepackage{pdfpages}
\usepackage{textcomp}
\usepackage{xcolor}
\usepackage{bm}
\usepackage{booktabs}
\usepackage{svg}
\svgsetup{inkscapelatex=false}
\usepackage{tikz}
\usepackage{float}
\usepackage{flushend}
\usepackage[hyphens,spaces,obeyspaces]{url}
\usepackage[]{hyperref} % optional

\def\BibTeX{{\rm B\kern-.05em{\sc i\kern-.025em b}\kern-.08em
    T\kern-.1667em\lower.7ex\hbox{E}\kern-.125emX}}
\usetikzlibrary{arrows,decorations.pathmorphing,backgrounds,fit,positioning,calc,shapes}

\usepackage[linesnumbered,ruled,algo2e,resetcount]{algorithm2e}%for the environment
\SetAlFnt{\small}
\SetAlCapFnt{\small}
\SetAlCapNameFnt{\small}

\begin{document}

%%%%%%%%%%%%%%%%%%%%%%%%%%%%%%%%%%%%%%%%%%%%%%%%%%%%%%%%%
% CASE LIMIT: 
% Regular paper: 6+2 pages (text+references)  DEADLINE: March 1st
% Industry Presentation: 4 pages (text+references) DEADLINE: April 1st
%%%%%%%%%%%%%%%%%%%%%%%%%%%%%%%%%%%%%%%%%%%%%%%%%%%%%%%%%

% ------ DECIDE ON A TITLE!
\title{\LARGE \bf
%Behavior Trees in Industrial Applications: A Case Study in Mining
%A Case Study of Behavior Trees: Underground Explosive Charging
Behavior Trees in Industrial Applications:\\ A Case Study in Underground Explosive Charging
\thanks{\textbf{This work has been submitted to the IEEE for possible publication.
Copyright may be transferred without notice, after which this version
may no longer be accessible.}}
}

%Authors metadata
\author{\authorblockN{
Mattias Hallen$^{a}$,
Matteo Iovino$^{b}$,
Shiva Sander-Tavallaey$^{b}$,
Christian Smith$^{c}$}
\thanks{$^{a}$ABB Mining R\&D, Umeå, Sweden}
\thanks{$^{b}$ABB Corporate Research, Västerås, Sweden}
\thanks{$^{c}$Division of Robotics, Perception and Learning, KTH - Royal Institute of Technology, Stockholm, Sweden}
}

\maketitle

\begin{abstract}
In industrial applications Finite State Machines  (FSMs) are often used to implement decision making policies for autonomous systems. In recent years, the use of Behavior Trees (BT) as an alternative policy representation has gained considerable attention. The benefits of using BTs over FSMs are modularity and reusability, enabling a system that is easy to extend and modify. However, there exists few published studies on successful implementations of BTs for industrial applications. This paper contributes with the lessons learned from implementing BTs in a complex industrial use case, where a robotic system assembles explosive charges and places them in holes on the rock face. The main result of the paper is that even if it is possible to model the entire system as a BT, combining BTs with FSMs can increase the readability and maintainability of the system. The benefit of such combination is remarked especially in the use case studied in this paper, where the full system cannot run autonomously but human supervision and feedback are needed.
\end{abstract}


\section{Introduction}

Deploying robots to automatize industrial applications increases efficiency, precision, and productivity, while reducing risks for human operators. Central to the effectiveness of the robotic system is the development and implementation of robust task switching policies that define the sequence of actions performed by the robots to solve the problem at hand.
The design of a policy for task planning is challenging because there exist several policy representation alternatives with different advantages and properties. This entails that depending on the use case one policy representation could be more suitable than another.

Some examples of task switching policy representations to control robot applications and industrial processes are Petri Nets, Finite State Machines (FSMs), Teleo-Reactive programs, and Decision Trees.

Another alternative that gained popularity in the past years as an alternative to FSMs is Behavior Trees (BTs)~\cite{colledanchise_behavior_2018}, especially for the properties of reactivity and modularity. Reactivity allows the system to pre-empt a running action to execute another with higher priority as a reaction to sudden events in the environment. Modularity enables the possibility to design and test every component, such as the robot actions, independently and then to combine and reuse them without breaking the logical correctness of the policy.

BTs are being widely used to implement AI in computer games and are recently becoming more common in robotics applications as well~\cite{iovino_survey_2022}. However, their usage is still mainly documented in efforts from academia while it is not yet clear to what extent they are applicable in industrial settings. For this reason, we share a successful implementation of BTs in an industrial scenario by providing useful technical suggestion on the BT design.

The use case scenario considered in this paper is automating the process of blasting charger in underground mines. For this application, the considered robotic system is responsible for automatically preparing the explosive charges and identifying the drilled holes in the rock walls where to insert the explosive.
The environment this system will encounter is highly unstructured and varied as every tunnel and every charging mission is unique. Thus, this application requires the system to be flexible and adaptable to changes in the environment. The system also requires a human operator to be in the loop for starting/stopping and modifying the currently active charging mission.
Although possible, defining a single BT to meet these requirements of the task would have led to several design challenges and usability limitations. For this reason, the task level policy responsible for controlling the robotic system is instead based on an high level FSM that allows the system to switch between control modes. Each of these control modes is in turn implemented as a BT that enables the robotic system to gather information about the environment and act on it.

Despite the individual prevalence of Finite State Machines (FSMs) and Behavior Trees (BTs)\----among the existing policy representations\----in various domains of robotics and automation, a combination of these methods remains relatively under-explored in academic research, with few pieces of evidence of its adoption within industrial contexts. By stating the reasoning behind the design choice together with the major advantages and shortcomings, this paper contributes with the following:
\begin{itemize}
    \item Describing the possibilities and challenges of implementing BTs on an industrial application.
    \item Describing the possibilities of combining FSMs and BTs to overcome the challenges that these representations entail when used independently. 
    \item Demonstrating the combination of FSMs and BTs on a real industrial problem in an unstructured and varied environment.
\end{itemize}

The remainder of this paper is organized as it follows. Section~\ref{Background and Related Work} provides a background on BTs and FSMs and how they are deployed in the existing literature. In Section~\ref{Problem statement} the domain specific problem is explored and detailed. Section~\ref{Proposed solution} presents the proposed solution of combining FSMs and BTs to control the robotic system to solve the task. Finally, Section~\ref{Outcome} reviews the outcome of this study and the challenges associated with using this combined design for industrial use cases. 

\section{Background and Related Work} \label{Background and Related Work}

\subsection{Behavior Trees}

Behavior Trees (BTs) are a versatile and modular task-switching paradigm initially developed in the gaming industry and later extended to robotics applications~\cite{colledanchise_behavior_2018}. Represented as directed trees, BTs employ a depth-first pre-order traversal for recursive ticking. At a given frequency (the ticking frequency is not fixed but can be defined at a design stage) the tick signal propagates from the root node and recursively from the left-most child to the right-most one. The BT structure consists of internal nodes (\emph{control nodes} represented by polygons in Figure~\ref{fig:bt}) which determine how the tick signal is propagated to the children and execution nodes (ovals in Figure~\ref{fig:bt}), or behaviors, that are implemented in the leaves of the BT and determine how the agent acts on the environment.

%The BT structure consists of internal nodes (\emph{control nodes} represented by polygons in Figure~\ref{fig:bt}) including \emph{Sequence} (executing children sequentially and returning once all succeed or one fails), \emph{Fallback} or \emph{Selector} (executing children in sequence but returning once one succeeds or all fail), and \emph{Parallel} (executing children in parallel, returning when a pre-determined subset of children is successful).

Control nodes are divided into \emph{Sequence} executing children sequentially and returning once all succeed or one fails, \emph{Fallback} or \emph{Selector} executing children in sequence but returning once one succeeds or all fail, and \emph{Parallel} executing children in parallel, returning when a pre-determined subset of children is successful.

Execution nodes are classified into Action nodes executing behaviors capable of returning \emph{Running}, \emph{Success}, or \emph{Failure} statuses and Condition nodes immediately returning \emph{Success} or \emph{Failure} based on status checks. BTs explicitly support task hierarchy, action sequencing, and reactivity~\cite{iovino_survey_2022}. The modular design allows for the seamless testing, interchangeability, and reusability of subtrees without compromising the overall structure~\cite{iovino_programming_2022, colledanchise_how_2017, biggar_modularity_2022}.

Unlike other policy representations, BTs employ a Running return state. This state facilitates prolonged execution of actions across multiple ticks and enables reactivity by pre-empting running actions for higher-priority behaviors. The standardized I/O structure, with a tick signal initiating execution and return statuses as the output for all behaviors, is the key features that enables modularity in BTs. This architecture positions BTs as a flexible and effective solution for task-switching controllers in both gaming and robotics domains.

\subsection{Finite State Machines}

Finite State Machines (FSMs) are derived from state automata and consist of a collection of states interconnected by transitions (see Figure~\ref{fig:fsm} as an example). Each state encapsulates a controller responsible for robot behavior, generating effects in the environment upon execution. The occurrence of these effects triggers an event, facilitating the transition to the subsequent state. FSMs, inclusive of Sequential Function Charts (SFCs)\---–a graphical programming language for Programmable Logic Controllers (PLCs)\----find extensive use in industry due to their intuitive design and implementation simplicity~\cite{lepuschitz_using_2019}.

However, FSMs present a challenging trade-off between reactivity and modularity. The execution of FSMs can be likened to the early programming languages' GoTo statement, where the execution flow jumps abruptly between different parts of the program~\cite{colledanchise_behavior_2018}. This GoTo-style execution, discouraged in programming and robotics, is a drawback of FSMs. %In contrast, Behavior Trees (BTs) execute akin to a function call, enabling a flow jump to another part of the code and subsequent return after completion.

The reactivity of FSMs necessitates numerous transitions, complicating maintenance when states are added or removed~\cite{iovino_programming_2022}. This renders FSMs less modular and scalable. Efforts to address modularity include grouping states into hierarchies, as seen in the design of Hierarchical State Machines (HFSMs)~\cite{colledanchise_how_2017} at the expenses of readability.

FSMs further exhibit the drawback of requiring manual intervention upon failure if explicit transitions to recovery states are not implemented.

\subsection{Related Work} \label{Related Work}

Because of the increasing interest and popularity of BTs, there exist several works comparing BTs with FSMs and other architectures. In particular, in~\cite{biggar_expressiveness_2021, biggar_modularity_2022, colledanchise_how_2017} BTs are proven to generalize other policy representations, such as FSMs, Petri Nets, Decision Trees, and the Teleo-Reactive programs. Furthermore, in~\cite{marzinotto_towards_2014, colledanchise_how_2017} it is shown how to build an FSM that behaves like a BT, trading off modularity and fault tolerance with readability. In~\cite{iovino_programming_2022} instead, BTs and FSMs are compared on a series of robotics tasks where FSMs become progressively harder to maintain and modify. However, the paper points out that with the choice of a fault-tolerant design of the FSM, the robot behaves the same as when controlled by a BT. 

From this body of work, it follows that since in FSMs states have access to internal variables and past decisions, they are more expressive than BTs~\cite{biggar_expressiveness_2021}.
Furthermore, BTs are more reactive than FSMs due to the fact that at every execution cycle, the tick signal propagates from the root down to the whole tree, thus re-evaluating conditional statements on the world and robot states. BTs are also more readable than FSMs as they do not encode past decisions in the representation~\cite{biggar_modularity_2022} and have optimal modularity due to the tree representation~\cite{biggar_modularity_2022, iovino_programming_2022}. Modularity is also a key enabler for automatically generating BTs given a task description as in~\cite{iovino_framework_2023, styrud_bebop_2023}.

Unlike FSMs, the modularity property is enforced in HFSMs since the states are grouped hierarchically~\cite{colledanchise_how_2017}. Thus, HFSMs offer a more organized solution where the transition from one operative mode to another is facilitated. At a lower level, these execution states can in turn be implemented with other policy representation. 
HFSMs were used as the task planner in the complex DARPA Robotics Challenge involving both navigation and manipulation for humanoid robots in unknown and unstructured environments~\cite{kohlbrecher_comprehensive_2016, schillinger_human-robot_2016, romay_collaborative_2017}. The peculiarity of this solution is that the HFSM enables semi-autonomous behaviors for the agent  where the human is kept in the loop.

However, designing the lower states behaviors similar to BTs opens up for hybrid approaches that can be quite efficient. For instance, in the study by Wuthier et al.~\cite{wuthier_productive_2021}, an FSM operates in sync with a BT, facilitating smoother transitions between skills, especially in cases of preemption where a higher-priority skill interrupts the current one. Authors in~\cite{zutell_flexible_2022, zutell_ros2_2022} highlight that while BTs lacks cyclic behaviors in their execution, the HFSMs are inherently equipped with this characteristic when implementing looping transitions. To address this, they propose a hybrid method wherein an HFSM serves as the top-level task planner, with its states represented as BTs. Similarly,~\cite{coronado_towards_2021} presents a comparable approach using FSMs instead, which is similar to the approach used in this paper. The combination of BTs with (H)FSMs enables user supervision and adjustable autonomy to best suit the different requirements of complex applications.

%\todo{ Based on this we conclude that we can use the combination of BT and FSM for better. Not as complex as HFSM then FSM.  Skapa egen Paragraf FSM och BT är enkla nog att kombinera. Det kommer vi demonstrera i det här paper. Det på platsen där referens 16 är i C. Related work. Cominbing FSM and BT and therefore demonstrate why they work together. Hitta om det finns någon som blandat detta.. isåfall i likhet med: annars: TO the best of our knowledge this is the first documented use:.}

\section{Description of the Use Case} \label{Problem statement}

%1. scanning of the wall with holes identification
%2. extraction of the casette with the detonator from the magasine
%3. extraction of the primer from the magasine
%4. assembly of the primer into the casette
%5. preparation for handover between robots
%6. insertion in the wall
%\todo{Execetuve summary. (Kopplpar tillbaka till related work).
%CHallenges först: 
%Börja med att sammanfatta vi ska visa ett use case som behöver det här.
%NU följer detaljerna.  }
%The drill and blast method commonly used in mining and tunneling operation is today highly mechanized. Mechanized equipment has moved operators from the rock face into vehicle cabins and lately into control rooms. However, the actual charging process still involves manual labor with operators hand-feeding a hose to fill explosives into the drilled holes.  

In the past century, underground mining has evolved from dangerous manual labour to a highly mechanized process. New technology and equipment such as haulers and loaders have removed the operator step-by-step from the unsecured rock face into a vehicle cabin and in recent years in some cases to the remote operating stations. However, this development has not reached the explosive charging process that remains at large a manual process involving humans at exposed rock faces.

Every tunnel section developed with the common drill and blast method is unique. Therefore, a system capable of charging rock faces needs to be reactive and adaptable to the uniqueness of the rocky environment. Furthermore, for safety reasons, the system cannot be fully autonomous but it should allow operators to provide input commands and to be able to pause and resume the operation in every step. Moreover, power outages and emergency stops will occur and it is thus important to give operators the tools to easily monitor and restart the system when required. The demand on operator supervision and adjustable reactive autonomy that is required by this use case prompted the combination of BTs and FSM described in Section~\ref{Related Work} as a suitable design choice for the task-level control of the robotic system.

\subsection{Application Example}

\begin{figure}[ht]
    \centering
    \includegraphics[width=.9\linewidth]{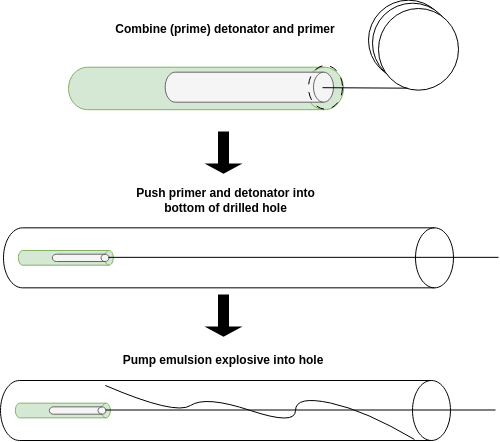}
    \caption{Workflow of priming explosives and pumping emulsion into a single drilled hole. This process repeats for every single drilled hole.}
    \label{fig:charging_workflow}
\end{figure}

At the site considered for this case study, the tunnel development charging utilizes three different explosive products in each drilled hole. The first is a detonator cap with attached cord utilized for initiating the blast. Thereafter a primer (explosive booster) charge responsible for initiating the primary explosive. Finally, the primary explosives is a pump-able emulsion. The detonator and primer are assembled by hand and placed into the tip of an extendable emulsion hose. Then, the hose is fed to the bottom of the charge hole and the emulsion is deposited while retracting the hose, leaving the primer and the detonator at the bottom while filling the hole with emulsion. This process is repeated for every blast hole that has been drilled in the rock face. The process is schematically summarized in Figure~\ref{fig:charging_workflow}. This process require dexterous hand movements from the operator to assemble detonator and primer. Furthermore, it is necessary to guide an emulsion hose while simultaneously handling the detonator cord.

This conventional method of charging requires operators to stand for extended periods of time close to the exposed rock face. This can be a dangerous part of the mine as it is a temporary site which will be blasted. By moving operators away from this area the safety of mining operations can be increased. Furthermore, charging is the only remaining manual labour in the drill and blast method. Automating this part is an important enabler to reach the desired autonomy within mining.

\subsection{Robotic System}

\begin{figure*}
    \centering
    \begin{subfigure}[c]{0.19\textwidth}
        \centering
        \includegraphics[width=\textwidth]{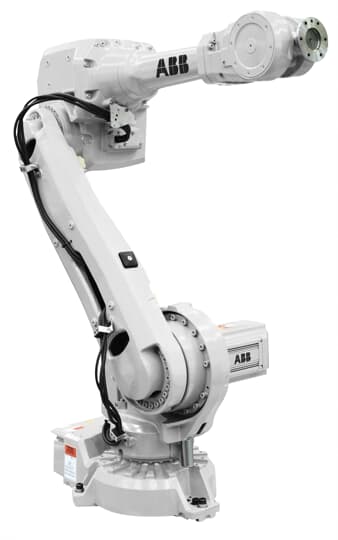}
        \caption{ABB IRB4600.}
        \label{fig:irb4600}
    \end{subfigure}
    \hfill
    \begin{subfigure}[c]{0.14\textwidth}
         \centering
         \includegraphics[width=\textwidth]{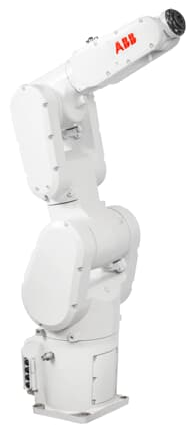}
         \caption{ABB IRB1300.}
       \label{fig:irb1300}
    \end{subfigure}
     \hfill
     \begin{subfigure}[c]{0.47\textwidth}
         \centering
         \includegraphics[width=\textwidth]{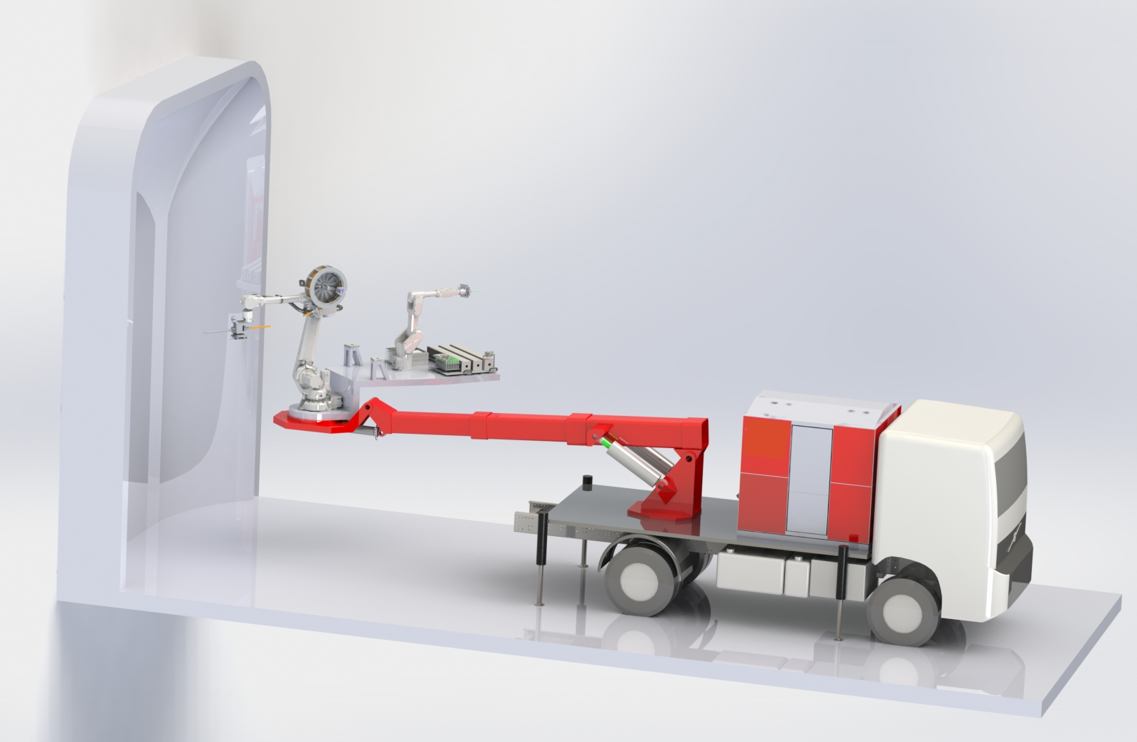}
         \caption{Full system used for the task.}
         \label{fig:step3}
     \end{subfigure}
    \caption{Robotic system for autonomous charge placement. The truck houses emulsion equipment and a hydraulic crane. This hydraulic crane extends the reach of the primary robot mounted at the end. On the robot platform the smaller secondary manipulator handles detonator and primer assembly.}
    \label{fig:robot}
\end{figure*}

The robotic system implemented for this task consists of two 6-DoF industrial manipulators (an ABB IRB4600 on the left of Figure~\ref{fig:robot} and an ABB IRB1300 on the right) placed on a 4-DoF hydraulic boom. The larger IRB4600 is referred to as the primary manipulator and it handles the positioning of the hose into the charge holes. The smaller IRB1300 is referred to as the secondary manipulator and it handles the reloading of primer and detonator explosives. Finally, the hydraulic boom supports the manipulators and extends their reach to cover the entire working area. A video showing an operational example of the system is publicly available\footnote{\url{https://www.youtube.com/watch?v=koyRJszHKBQ}}.

\subsection{Challenges} \label{sec:use_case_challenges}

The major challenge for the current robotic system is to work in a previously unknown environment. This places special demands on the robotic system, including that the system should be equipped with reliable computer vision algorithms to map the environment by identifying the charge holes and avoid collisions. In addition, the system should be reasonably fast in reacting and correcting its status depending on various unforeseen events.

Furthermore, the two manipulators have a parallel workflow which could be halted at any state. In case of a halted situation it is crucial that the execution is recovered without requiring a full reset of the system. 

The system also needs to be supervisable and respond to operator input. As a face can have up to 100 holes to be charged, the order of blasting is important, and if many holes are incorrectly charged the system needs guidance from an operator. This puts a requirement of flexibility on the system and re-adaption if problem occur during charging. As a first step, involving experienced operator input to plan and re-plan during the charging mission is important. % However, as data is collected of the charging process, some decisions could be taken by the system in the future.
However, as data on the task execution is being collected, it would be possible to automate some decision making processes in future deployments.

\section{Proposed solution} \label{Proposed solution}

To address the challenges and requirements for the task-level controller described in Section~\ref{sec:use_case_challenges}, we propose to design a high level FSM where each of the states is independently implemented as a BT.

The high-level FSM (shown in Figure~\ref{fig:fsm}) is responsible to transition the execution between the following operative modes:
\begin{enumerate}
    \item Scan the working area (obtain situational awareness and identify obstacles).
    \item Detect holes to charge.
    \item Plan the charging of the holes, which detonator to use and how much emulsion to inject.
    \item Charge the holes with the explosive prepared in the previous step.
\end{enumerate}

Having a clear separation of tasks facilitates the supervision and the input from the operators in the following circumstances:

\begin{itemize}
    \item \textit{Transitioning between operation modes.} In the charge planning stage the system automatically generates a plan to determine in which order the holes are charged. If for instance the operator detects some anomaly in the plan they can prompt the system to abort the mission and re-plan the charging sequence. 

    \item \textit{Obtaining operator inputs.} If we consider anomalies at a planning stage again, for instance in the event of a hole that failed to charge, the system could need operator guidance to continue the mission. The operator is then able to supply additional information to the system by for instance reposition the robot end-effector through teleoperation for a better alignment.

    \item \textit{Restarting tasks.} If the mission execution is halted, due to an emergency stop or a power outage, the system needs to be able to resume the task where it was left off. Through the high-level FSM the operator has the possibility to provide the system the correct input from where to resume the execution.
\end{itemize}

The above challenges could be addressed in a single BT by storing in the tree the execution state of the behaviors that were previously run. This feature is implemented by control nodes with memory. Introducing nodes with memory in a BT increases the type of switching strategies that the BT can express but at the cost of losing transparency as they implement logic that is hidden within the nodes and thus not possible to infer by just looking at the tree~\cite{biggar_expressiveness_2021, biggar_modularity_2022}. A BT design with memory nodes also impacts reactivity as it avoids the re-execution of some nodes and it is thus discouraged~\cite{colledanchise_behavior_2018}. 

Another consideration is that integrating the operator input into the BT would increase the size of the tree, increase the complexity of the structure and the dependencies between subtrees, and by consequence reduce the ability of the operators to follow the flow of the execution. Therefore, designing the whole policy as a BT would only be desirable if the system had to operate in full autonomy, thus without operator inputs or supervising production control systems. However, at the current state of the system every step of the mission is clearly separated from the others and inputs from operators are required to advance from a step to the next.
One example is the process of scanning the holes and planning the charge mission, which needs to be defined and initiated by the operator.

Thus, these missions steps are defined as states in an overarching FSM where they are then independently designed and executed as a BT. The benefits of this design structure are twofold. First, by wrapping the higher level functionality in an FSM it is possible to avoid introducing memory into the BT and simplify the overall architecture. This design decision was also driven by the fact that the reactiveness provided by a BT is not required at this high-level decision making. Second, the BTs are used where their key features are exploited at their full potential, such as where the system needs to be reactive and robust to failures that can happen at the execution stage.

To conclude, as an overview of the combined structure, the \textit{Charging} state is the most critical one where the most part of the execution is carried out. The other states can be considered as setup and information gathering steps that are necessary to generate the charging missions which is then executed by the BT.

\subsection{Finite State Machine}

\begin{figure}[H]
\centering
\def \globalscale {0.700000}
\begin{tikzpicture}[y=1cm, x=1cm, yscale=\globalscale,xscale=\globalscale, every node/.append style={scale=\globalscale}, inner sep=0pt, outer sep=0pt]
  \path[fill=white] ;
  \path[draw=black,fill=white] (2.4606, 9.816) ellipse (1.0583cm and 1.0583cm);
  \begin{scope}[shift={(-0.0132, 0.0132)}]
    \node[text=black,anchor=south] (text3080) at (2.4606, 9.7102){Pre-scan};
  \end{scope}
  \path[draw=black,fill=white] (2.4871, 7.1702) ellipse (1.0583cm and 1.0583cm);
  \begin{scope}[shift={(-0.0132, 0.0132)}]
    \node[text=black,anchor=south] (text6849) at (2.4871, 7.0644){Detect Holes};
  \end{scope}
  \path[draw=black,fill=white] (2.4871, 3.9952) ellipse (1.0583cm and 1.0583cm);
  \begin{scope}[shift={(-0.0132, 0.0132)}]
    \node[text=black,anchor=south] (text7437) at (2.4871, 3.8894){Charge plan};
  \end{scope}
  \path[draw=black,fill=white] (2.4871, 1.0848) ellipse (1.0583cm and 1.0583cm);
  \begin{scope}[shift={(-0.0132, 0.0132)}]
    \node[text=black,anchor=south] (text770) at (2.4871, 0.979){Charging};
  \end{scope}
  \path[draw=black,miter limit=10.0] (1.4023, 1.3494) .. controls (0.344, 2.2313) and (0.3094, 3.0776) .. (1.2986, 3.8883);
  \path[draw=black,fill=black,miter limit=10.0] (1.406, 3.9764) -- (1.3213, 3.7875) -- (1.2986, 3.8883) -- (1.2039, 3.9306) -- cycle;
  \begin{scope}[shift={(-0.0132, 0.0132)}]
    \node[text=black,anchor=south,fill=white] (text2741) at (0.7027, 2.3283){Re-plan};
  \end{scope}
  \path[draw=black,miter limit=10.0] (2.4871, 2.9369) -- (2.4871, 2.3117);
  \path[draw=black,fill=black,miter limit=10.0] (2.4871, 2.1728) -- (2.3945, 2.358) -- (2.4871, 2.3117) -- (2.5797, 2.358) -- cycle;
  \begin{scope}[shift={(-0.0132, 0.0132)}]
    \node[text=black,anchor=south,fill=white] (text5239) at (2.5135, 2.5135){Start charging};
  \end{scope}
  \path[draw=black,miter limit=10.0] (2.4871, 6.1119) -- (2.4871, 5.2221);
  \path[draw=black,fill=black,miter limit=10.0] (2.4871, 5.0832) -- (2.3945, 5.2684) -- (2.4871, 5.2221) -- (2.5797, 5.2684) -- cycle;
  \begin{scope}[shift={(-0.0132, 0.0132)}]
    \node[text=black,anchor=south,fill=white] (text7691) at (2.3, 5.4769){New holes detected};
  \end{scope}
  \path[draw=black,miter limit=10.0] (2.4606, 8.7577) -- (2.4786, 8.3968);
  \path[draw=black,fill=black,miter limit=10.0] (2.4855, 8.2582) -- (2.3839, 8.4384) -- (2.4786, 8.3968) -- (2.5688, 8.4476) -- cycle;
  \path[draw=black,miter limit=10.0] (3.519, 4.2598) .. controls (4.0481, 4.9653) and (4.3127, 5.6268) .. (4.3127, 6.2442) .. controls (4.3127, 6.8615) and (4.1035, 7.2121) .. (3.6851, 7.2959);
  \path[draw=black,fill=black,miter limit=10.0] (3.5491, 7.3231) -- (3.7489, 7.3776) -- (3.6851, 7.2959) -- (3.7124, 7.1959) -- cycle;
  \begin{scope}[shift={(-0.0132, 0.0132)}]
    \node[text=black,anchor=south,fill=white] (text7599) at (4.32, 5.9267){Scan again};
  \end{scope}
\end{tikzpicture}
    \caption{Scheme of the state machine. Each state has a corresponding Behavior Tree that triggers upon entering the state.}
    \label{fig:fsm}
\end{figure}
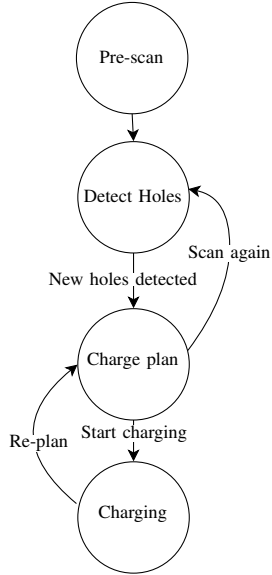

The state machine follows the high level task list described in Section~\ref{Proposed solution}. This simple FSM follows the task list with the important possible restarts highlighted. Each state has a corresponding BT which executes upon transitioning into that state. A description can be seen in Figure~\ref{fig:fsm}.

The most complex BT is found in the \textit{Charging} state. Given a charging mission which is supplied by the operator in the \textit{Charge Plan} state, this tree executes the full charging cycle. If failures happen during the execution, they are notified in the FSM which then requests an appropriate input from the operator. The other BTs are relatively simpler and they are thus omitted in the interest of space.

\begin{figure*}[ht]
    \centering
    \includegraphics[width=\linewidth]{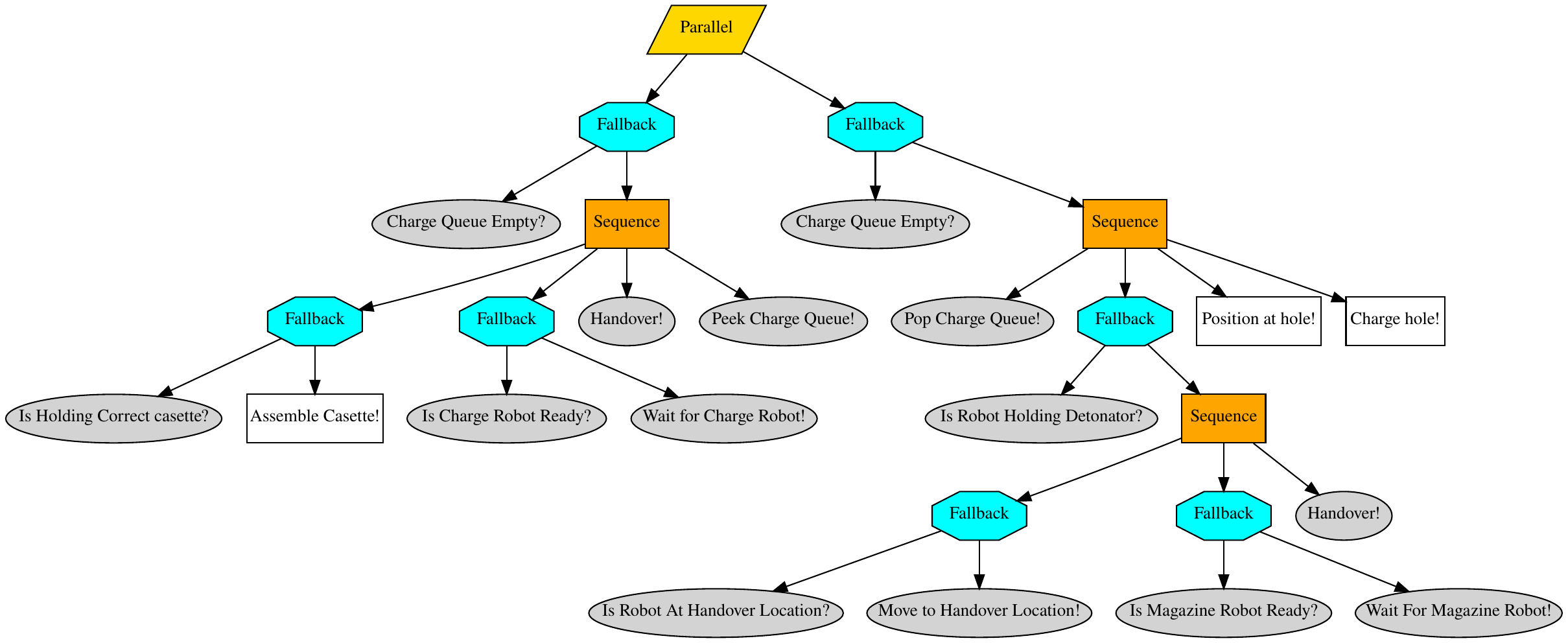}
    \caption{Scheme of the Charging BT. For simplicity, leaves or fallback behaviors that are not useful to understand the functioning principle of the BT were omitted. The white boxes with a label represent subtrees achieving a subtask described by the label.}
    \label{fig:bt}
\end{figure*}

\subsection{Behavior Tree} \label{tree structure}

The BT is implemented using the \texttt{BehaviorTree.CPP} library and developed with the BT user interface \texttt{Groot}~\cite{mood2be}. This was integrated into ROS2~\cite{ros} where each BT is hosted by an action server expecting an input goal and a feedback reply on the execution result. The execution result and feedback is directly tied to the result of the BT.

The charging BT is reported in Figure~\ref{fig:bt} and it is considered the most complex as it contains the parallel workflow of both manipulators as well as movements of the boom. This BT is designed with a parallel root node with two children, one controlling the robot handling explosives and the other one controlling both the charging robot and hydraulic boom. The parallel root node has a \textit{``success on all''} policy which makes the BT return Success when all the children are successful.
The children are designed with the backward-chaining design principle~\cite{colledanchise_towards_2019, gustavsson_combining_2022} where conditions are connected under a Fallback node with actions that achieve them. The cyclic task of charging several holes is modeled with a queue popping mechanism with an explicit goal to empty the queue of charge holes contained in the charging mission.

The primary manipulator controls the popping operation of the queue. This removes the first charge hole from the mission and put the information of this charge hole on the blackboard. The two manipulators are synchronized in the handover skill which involves the handover of explosives from one manipulator to the other. After the handover is performed, the manipulator handling explosives peeks the next charge hole on the queue to initiate preparation while the charging manipulator performs the hose insertion.

\section{Outcome} \label{Outcome}
\begin{figure*}[!ht]
    \centering
     \begin{subfigure}[b]{0.40\textwidth}
         \centering
         \includegraphics[width=\textwidth]{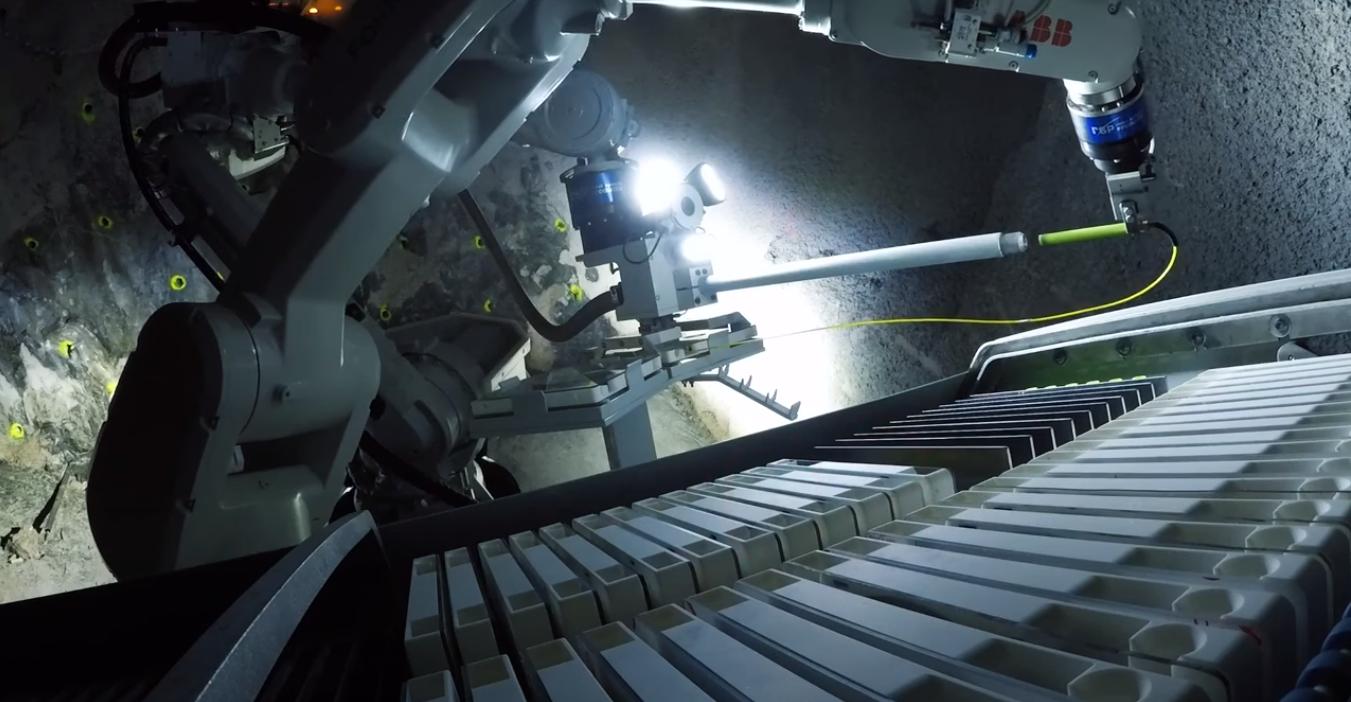}
         \caption{Handover between the robots.}
         \label{fig:step5}
     \end{subfigure}
     % \hfill
     \qquad \qquad
     \begin{subfigure}[b]{0.40\textwidth}
         \centering
         \includegraphics[width=\textwidth]{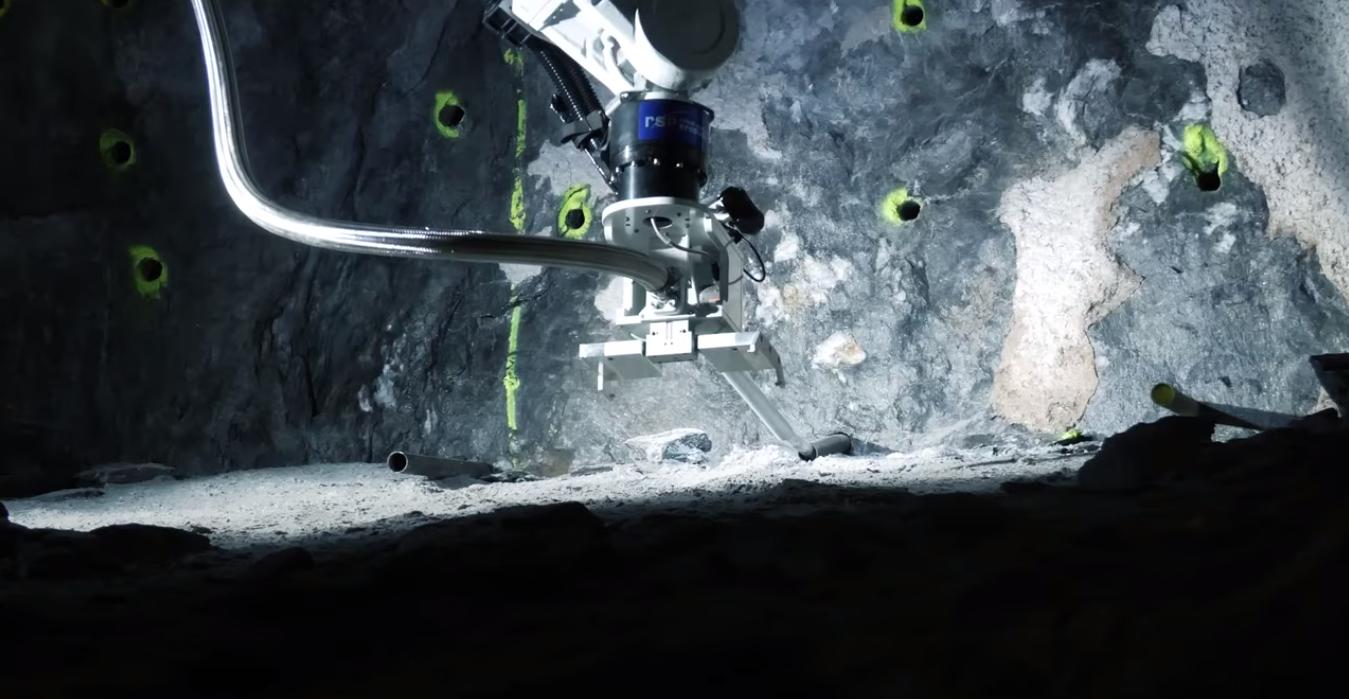}
         \caption{Charging of one hole.}
         \label{fig:step6}
     \end{subfigure}
        \caption{Images from field tests during 2023.}
        \label{fig:process}
\end{figure*}

The system was successfully demonstrated in a real underground mine in 2023\footnote{\url{https://new.abb.com/news/detail/108967/abb-boliden-and-lkab-complete-successful-testing-of-industry-first-automated-robot-charger-for-increased-safety-in-underground-mines}}. During the demonstration, the full workflow was executed including the explosive charging in the rock face. Some key steps of the demonstration are reported in Figure~\ref{fig:process}.

\subsection{Lessons Learned}
The higher level task switching remained constant throughout development. However, the BTs were constantly modified and improved with the progress of the complexity. Thanks to the modularity feature of BT, adapting the behaviors to the new challenges required little effort.

One strength of BTs is the possibility to add local recovery behaviors by reusing skills that are implemented for the general task execution. Some examples are: 
\begin{itemize}
    \item 
    Before approaching a hole to charge, the BT checks if the robot is currently holding a detonator. If that is the case it proceeds to charge the hole otherwise it retrieves a new detonator. This behavior is implemented by the subtree rooted with a Fallback node and goal condition \texttt{\small Is Robot Holding Detonator?} in the right-hand side of Figure \ref{fig:bt}.
    \item
    At the event of failing to find a charge hole within \texttt{\small Position at Hole!}, the robot can sweep a small area to try to find the hole with a different viewing angle.
    \item
    In the \texttt{\small Charge hole!} if the hose feed does not succeed, various strategies such as wiggling the hose and insertion/retraction are implemented to try to get past any eventual blockage.
    \end{itemize}

These are some examples of behaviors that were added after the initial design of the system to face the complex facets of the task. With only re-using existing skills, complex behaviors can be implemented by modifying the existing BT with little effort, as shown in~\cite{iovino_programming_2022}.

Conditions that require operator assistance are also easily implemented into the same architecture. A failure status would propagate upwards the tree and thus signaling the FSM for operator assistance.

However, even though it is tempting to define all behaviors at an atomic level to increase their chances of re-usability, this goes at the expenses of the BT size and the readability.
For instance, although it is possible to define action behaviors to control the motors of each robot joint, the benefits do not justify the increased complexity and it would be simpler to integrate the joints motion into one single robot motion behavior instead.

\subsection{Faced Challenges} \label{Challenges of the industrial applications}
%\todo{More challenges, e.g. byggstenarna för BT fanns ej. Best-practice examples, t.ex. backwards chain, loops. Easy fall into sequence with memory -- Not only toolbox also mindset.}
%\todo{look into parallel execution in FSMs}

%One of the main challenges with implementing BTs for this application is execution without a predefined goal, where the system requires input from human operators. While practically possible, the visualization of where the system requires input becomes hard to read when combined with other task execution subtrees. By combining BT with a higher level FSM for task switching the BT could be kept cleaner and thus increased the readability.
One of the main challenges faced when deciding which policy representation to use, was to find a design that would allow the system to solve most of the task in full autonomy but still allowing the operators to intervene and change the mission details. By modelling the full system with a BT, it would have been hard to distinguish the action nodes controlling the manipulators from those prompting the operators for inputs. Designing a higher-level FSM to handle the interaction with the operator, makes the underlying BT cleaner and simpler to read.

%Another challenge was the lack of \emph{Execution nodes} readily available. These had to be created from scratch, but by integrating generic ROS Action and Service Calls into BT Action and Condition nodes this could be solved outside of the BT implementation. But having a palette of nodes that solves the most general problems would be incredibly helpful.
Due to the uniqueness of the mission and the lack of published work on industrial applications of BTs, another challenge was to create all the \emph{Execution nodes} from scratch. We decided to implement the actions and condition nodes in the BT as ROS Actions and Services respectively. In this way, the structure of the BT node is disentangled from the actual ROS low-level implementation, thus making the design more modular. 

The lack of published best practices on industrial problems utilizing BTs is also an important factor. One common pitfall when designing BTs is to keep the mindset of designing FSMs. For instance, the \emph{Sequence} node in a BT is reactive, meaning that upon receiving a new tick, the node executes the left-most child firstly. Designing behaviors that only utilize non reactive sequences that remember the return statuses of the children (also known in literature as Sequence with memory~\cite{colledanchise_behavior_2018}) can diminish the benefits of using BTs. 

Another challenge is to model a cyclic execution in BTs. In this instance we use a queue popping mechanism which is effective but hard to read in the BT itself. This problem stems from the fact that the BT controls manipulators operating simultaneously and requiring synchronization. One alternative to explore is to design an individual BT for each of the two manipulators. In this case, the queue popping mechanism could be implemented with dedicated ROS Services that provide the location or the ID of the next hole to charge.

\subsection{Concluding Remarks} 
%\todo{expandera}

In this study the we have utilized BTs to control a robotic system for automated charging in a mining application. We have faced several challenges, many of them related to the lack of publications on the implementation of BTs for industrial applications and on best design practices for BTs.

%Despite the challenges, the modularity achieved by BT and by forcing the implementer view the set of actions the system can take in a set of skills, and by limiting the implementer to specify it as a success or a failure the implementer ends up with a toolbox of composable behaviors to build the complex logic of the system. Which has been helpful in this study to rapidly implement new behaviors that were not foreseen to be necessary before field trials commenced.
Using BTs forces the designer to organize the set of skills that are available to the system in a set of action nodes with a pre-defined structure: the tick signal as input to trigger the execution and the return statuses as output. With this interface, the designer can quickly define a toolbox of composable behaviors to build complex logic for the system. For instance, this allowed us to rapidly implement behaviors to handle failure cases that were not possible to predict before the field trials.

%Although not an inherit property of BT, the implementation of the framework is also a substantial factor in the experience of developing a system relying on BTs. The use of \texttt{BehaviorTree.CPP} library and with the aid of a graphical user interface \texttt{Groot}~\cite{mood2be} was an important factor in the successful implementation.
Another remark to make is on the choice of the BT development environment and tools. In our case, using the \texttt{BehaviorTree.CPP} library and with the aid of the graphical user interface \texttt{Groot} greatly facilitated the implementation.

Even though BT generalize other structures~\cite{colledanchise_how_2017}, implementing too many functionalities of the system into a single BT can cause problems in terms of readability and maintainability. Instead, by utilizing the FSM for switching between operating modes and keeping the BTs for lower level reactive behaviors it is possible to design a system that exploits the benefits of BTs and minimizes the drawbacks.

Finally, BTs have received considerable attention in robotics applications within academia~\cite{iovino_survey_2022}. This study provides one example of a successful implementation in an industrial scenario by exploring the applicability of BTs within industry and by providing technical suggestions on BT design practices that can be utilized for other systems.

%\section*{Acknowledgments}

\bibliographystyle{IEEEtran}
\bibliography{references.bib}

\end{document}